\documentclass{article}

\usepackage{microtype}
\usepackage{graphicx}
\usepackage{booktabs} 
\usepackage{amsmath}
\usepackage{soul}
\usepackage{booktabs}
\usepackage{algorithm}
\usepackage{algorithmic}
\usepackage{subcaption}
\usepackage{amsfonts}
\usepackage{multirow}

\usepackage{hyperref}

\usepackage[accepted]{icml2021}

\icmltitlerunning{Y-Drop: A Conductance based Dropout for fully connected layers}

\begin{document}

\twocolumn[
\icmltitle{Y-Drop: A Conductance based Dropout for fully connected layers}





\begin{icmlauthorlist}
\icmlauthor{Efthymios Georgiou}{to,goo}
\icmlauthor{Georgios Paraskevopoulos}{to,goo}
\icmlauthor{Alexandros Potamianos}{to,ed}
\end{icmlauthorlist}

\icmlaffiliation{to}{School of ECE, National Technical University of Athens, Athens, Greece}
\icmlaffiliation{goo}{ILSP, Athena RC, Athens, Greece}
\icmlaffiliation{ed}{Behavioral Signal Technologies, Los Angeles, CA, USA}

\icmlcorrespondingauthor{Efthymios Georgiou}{efthygeo@mail.ntua.gr}

\icmlkeywords{Dropout, Conductance, Regularization}

\vskip 0.3in
]



\printAffiliationsAndNotice{}  

\begin{abstract}

In this work, we introduce Y-Drop, a regularization method that biases the dropout algorithm towards dropping more important neurons with higher probability.
The backbone of our approach is neuron conductance, an interpretable measure of neuron importance that calculates the contribution of each neuron towards the end-to-end mapping of the network.
We investigate the impact of the uniform dropout selection criterion on performance by assigning higher dropout probability to the more important units.
We show that forcing the network to solve the task at hand in the absence of its important units yields a strong regularization effect. 
Further analysis indicates that Y-Drop yields solutions where more neurons are important, i.e have high conductance, and yields robust networks.
In our experiments we show that the regularization effect of Y-Drop scales better than vanilla dropout w.r.t. the architecture size and consistently yields superior performance over multiple datasets and architecture combinations, with little tuning.

\end{abstract}

\section{Introduction}

Neural Networks in the deep learning era tend to utilize up to billions of trainable parameters.
This creates a need for efficient regularization methods. 
Dropout \cite{hinton_improving_2012, srivastava_dropout_2014} is the most widespread regularization method for deep neural networks (DNNs), due to its simplicity and effectiveness \cite{krizhevsky2012imagenet, he2016deep}. 
The original algorithm proposes to randomly omit\footnote{We also refer to this procedure as \textit{dropping} units. We note here that dropping refers to the training stage alone and not the inference step. Literature also refers to dropping as applying a binary mask to the activations, i.e. masking out.} a portion of units during the forward and backward pass of the training procedure.
Despite the benefits of its probabilistic nature, dropout fails to 
capture important problem-specific characteristics related to the data and task at hand. 
This limitation raises questions regarding the optimality of dropout's selection criterion. 

\par
Motivated by these observations, dropout variants have been proposed to improve the original algorithm by embedding external knowledge.
One line of work includes 
approaches that take advantage of architectural or data-specific properties, \textit{e.g.} image locality \cite{devries_improved_2017}.
Another research direction studies \textit{heuristic} variants,
such as CorrDrop \cite{zeng_corrdrop_2020} which uses a feature correlation map to drop the least informative regions.
However, none of these approaches share the wide adoption of the original algorithm.
This is either due to the need for extensive tuning of the proposed methods, or to the task-specific nature and ad-hoc implementations of the proposed algorithms.





\par

Dropout also draws analogies with neuro-scientific studies. 
In particular \citet{mcdonnell2011benefits} discuss the benefits of noise in neural brain systems and \citet{montijn2016population} the robustness of such systems to noise.
In the machine learning context, training with noise \cite{sietsma1991creating, bishop1995training} has been shown to yield regularized solutions.
Dropout can also  be interpreted as a form of training with noise \cite{srivastava_dropout_2014}, which relies on randomness to prevent feature co-adaptation. 
However, one could directly try to battle co-adaptation.
Co-adaptation often manifests itself when a ``strong" neuron ``dominates" the contribution of a ``weak" neuron, i.e. when a neuron is only helpful in the presence of other specific neurons.
Motivated by this observation researchers have tried to find those strong neurons and drop them
to allow for weak neurons to train.
Guided Dropout \cite{keshari2019guided} utilizes a matrix decomposition heuristic to track strong neurons and drop them. InfoDrop \cite{shi_informative_2020} is used in computer vision tasks and drops units which are biased towards texture information. 


\par
In this work we integrate an importance attribution algorithm during the training procedure. 
We rely on Conductance \cite{dhamdhere2018important}, which is a measure of neuron importance
and calculates the contribution of each unit to the end-to-end mapping of the network.
We introduce \textit{Y-Drop} \footnote{Y is the symbol for conductance in circuit theory.} as a regularization approach, which augments vanilla dropout, by modifying each neurons drop probability based on its conductance score.
Intuitively, a regularizer should penalize 
network behavior that is associated with overfitting, such as large weight values or in our case the presence of limited important units.
We show that forcing the network to solve the task at hand in the absence of its important units results to strong regularization.
Moreover, neuron importance can be measured in a task-agnostic manner, i.e. without  modifications for different input modalities and architectural choices.
Therefore Y-Drop can be easily adapted to new scenarios.
\par
Our key contributions are:
1) we propose a novel extendable framework which integrates importance measure information with dropout, aiming for an interpretable approach,
2) we show that injecting conductance information during training, improves network generalization and scales with the architecture size while requiring little tuning,
3) we find that networks trained using Y-Drop utilize their capacity more efficiently, allowing more units to participate in solving the task at hand. 
Our code is available as open source\footnote{Link will be provided upon end of double blind period.}.

\section{Related Work}

Since the introduction of the dropout algorithm \cite{hinton_improving_2012}, several variants have been proposed. 
\textit{StandOut} \cite{ba_adaptive_2013} overlays a binary belief network on top of the neural network, in order to adaptively tune the dropout probability of every neuron.  
\textit{Variational Dropout} \cite{kingma_variational_2015} interprets dropout with Gaussian noise as maximizing a particular variational objective where dropout rates are learned.
\citet{gal2017concrete} propose \textit{Concrete Dropout} where the binary dropout masks are relaxed into continuous and the dropout probability is adapted via a principled optimization objective. 
Another variant is \textit{DropConnect} \cite{wan2013regularization} which randomly drops connections instead of activations. Other approaches like \textit{Annealed Dropout} \cite{rennie2014annealed} and \textit{Curriculum Dropout} \cite{morerio2017curriculum} propose dropout rate scheduling schemes.
\textit{JumpOut} \cite{wang2019jumpout} proposes a series of heuristics which can be integrated with dropout in fully connected layers and convolutional maps, e.g. sample the dropout hyperparameter from a monotonically decreasing distribution in every step.
\par
Other variants aim to exploit data or architecture specific properties and are mostly applied in Convolutional Neural Networks (CNNs) and Recurrent Neural Networks (RNNs). 
In CNNs \textit{DropPath} \cite{larsson_fractalnet_2017} zeroes out an entire layer during training and \textit{SpatialDropout} \cite{tompson_efficient_2015} stochastically drops an entire channel from a feature map.
For RNNs, \citet{pham2014dropout} propose to apply dropout only to the non-recurrent connections and \textit{ZoneOut} \cite{krueger_zoneout_2017} uses an ``identity" mask which is shared through time, thus preserving hidden activations rather than dropping them. Both approaches respect the weight sharing property of RNNs and allow gradients to flow during back-propagation.
CutOut \cite{devries_improved_2017} zeroes out parts of the input images. \citet{park_specaugment_2019} propose \textit{SpecAugment} which extends CutOut for spectrorgrams, by zeroing out frequency bands or consecutive time steps. 
In \textit{DropBlock} \cite{ghiasi_dropblock_2018}, authors propose to remove contiguous regions of a feature maps, exploiting the locality properties in image-based CNNs.
\par
Another line of research utilize a measure which encodes some metric about individual neurons or a region of the network.
For instance, \textit{CorrDrop} \cite{zeng_corrdrop_2020} exploits a feature map correlation heuristic and masks out those regions with small feature correlation, i.e with less discriminative imformation.
In the same spirit, \textit{InfoDrop} \cite{shi_informative_2020} uses an info-theoretic measure in order to merely preserve shape information and discard texture in CNNs. 
\textit{Guided Dropout} \cite{keshari2019guided}, uses a weight matrix decomposition heuristic in order to calculate the so called strength of every node.
Based on this heuristic it drops units by sampling from the pool of strong nodes alone.
Our approach also exploits a measure and in particular a neuron importance evaluation algorithm.
\par
The works regarding model interpretability (attribution algorithms),
evaluate the contribution of each input feature or neuron to the end-to-end mapping of the model. 
\textit{Integrated Gradients} \cite{sundararajan2017axiomatic} attribute the network's output to its input features. Specifically, given a sample (e.g image) and a baseline input (e.g a black image) they interpolate multiple samples along the line path which departs from the baseline and ends to that particular sample.\
The overall contribution is calculated as integrating over this line path of interpolated samples. 
\textit{Conductance} \cite{dhamdhere2018important, shrikumar2018computationally}, which we use in this work, is a measure of neuron importance which
calculates an importance score and is based on Integrated Gradients.
In particular, the conductance of any unit is equivalent to the flow of integrated gradients through this unit.
Other relevant approaches are
\textit{Gradient SHAP} \cite{gradient_shap_2017}, which is a gradient based method which assigns each feature an importance value for a particular prediction. Moreover, \textit{DeepLIFT} \cite{shrikumar2017learning} which is a back-propagation based method for input feature attribution. Also \textit{Internal Influence} \cite{leino2018influence} and \textit{GradCAM} \cite{selvaraju2017gradcam} attribute the output of the network to a given layer. 

\section{Background}
In this section we will revisit how neural conductance is calculated. We will also revisit the vanilla dropout formulation and introduce some useful notation.
\vspace{5pt}
\noindent
\textbf{Notation:}
For the rest of this work we refer to the network mapping as $y=\mathcal{F}(x ; \theta)$, where $y$ describes the network output, $x$ the corresponding input and $\theta$ denotes the collection of trainable parameters.
We omit the trainable parameter symbol when able to reduce notation.
For a fully connected architecture with $L$ layers, we set $l=\{1,\cdots, L\}$ as the fully connected layer indicator and $k=\{1,\cdots, N_l\}$ as the neuron index of the $l$-th layer. 

\begin{figure*}[t]
    \centering
    \includegraphics[width=\textwidth]{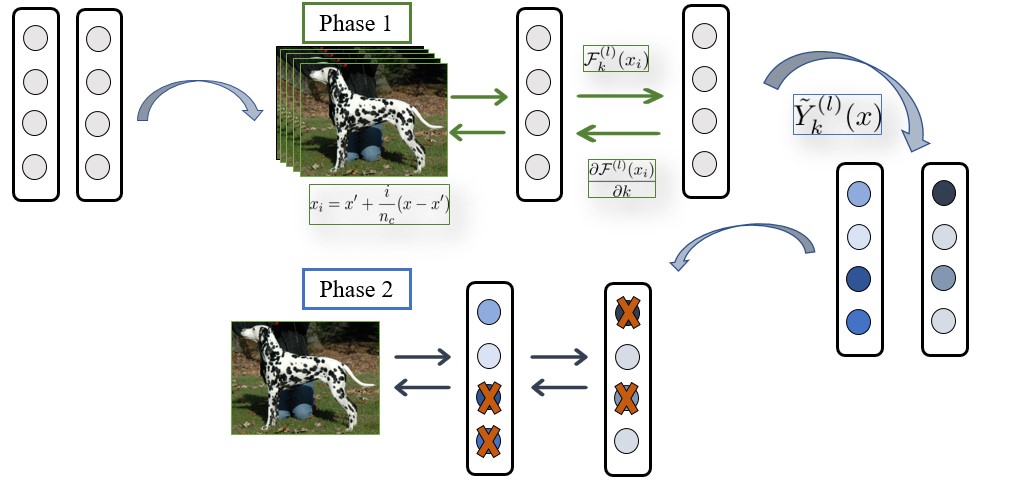}
    \caption{
    Y-Drop consists of two phases during each training step, conductance calculation and network update.
    To calculate conductance, we first interpolate samples over a given sample and an uninformative sample (e.g. a black image) and feed them to the network.
    For every unit in each layer, conductance is calculated based on the unit's activations (green forward pass) and the unit's partial derivatives (green backward pass) for all interpolated samples. 
    Darker colors denote units with higher per-layer conductance. 
    During the second phase, we use conductance scores for each unit to determine the unit's drop probability and the network parameters are updated through backpropagation.
    The curved arrows denote the transitions between phases.}
    \label{fig:Y-Drop}
\end{figure*}

\subsection{Conductance}\label{sec:conductance}
Conductance \cite{dhamdhere2018important} calculates the importance of each unit in the prediction of the network for a given input. For a given input $x$ and a reference baseline $x'$, e.g. a black image, conductance is defined as an integral over the line path from $x'$ to $x$. 
\footnote{Conductance is an attribution method that relies on perturbing the input to measure the change in predictions based on the perturbations. The baseline helps define these perturbations. \cite{dhamdhere2018important}}
Formally, the conductance of neuron $k$ in layer $l$ is defined as follows \cite{shrikumar2018computationally}:
\begin{equation}\label{cond_1}
    Y^{(l)}_k(x) = \int_{a=0}^{a=1} \frac{\partial \mathcal{F}(x' + a(x-x'))}{\partial \gamma_k^{(l)}(a)}\frac{\partial \gamma_k^{(l)}(a)}{\partial a} \mathrm{d}a
\end{equation}
where $\gamma_k^{(l)} (a)$ is the activation of neuron $k$ in layer $l$ given $x' + \alpha(x-x')$ as input vector and $\frac{\partial \gamma_k^{(l)}(a)}{\partial a} \mathrm{d}a$ denotes an infinitesimal step along the line path for unit $k$.
A more efficient reformulation, in terms of computational complexity, of Eq.~\eqref{cond_1} is \cite{shrikumar2018computationally}:

\begin{equation}\label{cond_2}
    Y^{(l)}_k = \sum_{i=1}^{n_c} \frac{\partial \mathcal{F}^{(l)}(x_i)}{\partial k} (\mathcal{F}_k^{(l)}(x_i)) - \mathcal{F}_k^{(l)}(x_{i-1}))
\end{equation}
where the integration is writen as a sum over discrete intermediate samples along the line path. $\mathcal{F}_k^{(l)}(x)$ is the activation of neuron $k$ in layer $l$ given input $x$. The interpolation points $x_i$ are calculated as of $x' + \frac{i}{n_c}(x - x')$. Naturally $n_c$ is the number of interpolation (or integration) steps. 
\par
The computational benefits of Eq. \ref{cond_2} are evident if we think that it can be computed by a single back-prop for all the intermediate interpolation steps, if we feed them as a batch of examples. 
We also note that this calculation is not computationally equivalent to a gradient calculation step with weight updates since the expensive calculation is in general the update of the weights.
For the rest of the paper we refer to the conductance of a unit $k$ in a fully connected layer $l$ as $Y_k^{(l)}$. 

\subsection{Dropout}\label{sec:dropout}
Dropout \cite{hinton_improving_2012, srivastava_dropout_2014} is a regularization method which injects noise in the training procedure and aims at preventing feature co-adaptation.
Dropout regularization is performed by randomly setting activations to zero during training.
In practice this is implemented as sampling a binary value, i.e. mask value, from a Bernoulli distribution.
Formally we describe this procedure as $m \sim Be(p)$, where $p$ is the probability of sampling a zero mask and is known as dropout probability or rate.
\par
After sampling the binary mask $m$, dropout algorithm employs a re-scaling trick which is expressed as:
\begin{equation}\label{rescale_trick}
    \tilde{m} = \frac{m}{1 - p}
\end{equation}
The re-scaled mask $\tilde{m}$ is then applied on the activations during the forward and backward pass of the training procedure.
The described process is repeated in every training step. 
During inference the dropout is ``deactivated", meaning that all units participate in the prediction of the network.

\section{Proposed Method}

The proposed method is illustrated in Fig.~\ref{fig:Y-Drop}. 
We apply the algorithm only to fully connected layers.
Each training step consists of two phases. During the first phase, we perform a forward and backward pass on interpolated samples to calculate conductance scores for each neuron. During the second phase, we use the conductance scores to drop the most important neurons in every layer and update the network parameters using backpropagation.
\subsection{Y-Drop} \label{sec:Y-Drop}

During the first phase of Y-Drop we calculate the conductance of each unit for every given training sample.
Specifically, a batch of interpolated images is fed to the network and then through a backward step we calculate the  necessary partial derivatives for Eq.~\eqref{cond_2}. This calculation is depicted with \textit{Phase 1} in Fig.~\ref{fig:Y-Drop}. 
Following Eq.~\eqref{cond_2} the conductance scores are calculated for every input sample.
However, since training is performed in mini-batches we need to calculate a single importance value for every neuron.
To achieve this we introduce the \textit{mean conductance per unit}, which is described in Eq.~\eqref{mean-importance}:
\begin{equation} \label{mean-importance}
    \tilde{Y}_k^{(l)} = \frac{1}{B_c} \sum_{j=1}^{B_c} Y_{kj}^{(l)} 
\end{equation}
where $B_c$ is the number of samples used for conductance calculation, and $Y_{kj}^{(l)}$ is the conductance score of neuron $k$ in layer $l$ for the $j$-th sample. 
Based on mean conductance, we rank the units in every single layer $l$ and consecutively separate them into two buckets, namely the strong $\mathcal{B}_{\mathcal{S}}^{(l)}$ and the weak $\mathcal{B}_{\mathcal{W}}^{(l)}$.
\par
In the second phase Y-Drop assigns higher drop probabilities to units with high mean conductance values.
The strong bucket, is assigned a high drop probability $p_{H}$, while the weak bucket, a low drop probability $p_{L}$. The bucket sizes are defined as  $w_{\mathcal{S}}, w_{\mathcal{W}} \in [0,1]$ for the strong and weak bucket respectively. We note that the bucket sizes should also satisfy $w_{\mathcal{S}} + w_{\mathcal{W}} = 1$. 
The operation of ``bucketization" is described via a mapping 
\begin{equation} \label{bucketization}
    g_{\mathcal{B}}(\tilde{Y}^{(l)}) = 
    \begin{cases}
    p_L , \   k \in \mathcal{B}_{\mathcal{W}}^{(l)} \\
    p_H , \   k \in \mathcal{B}_{\mathcal{S}}^{(l)}
    \end{cases}
\end{equation}


which is defined as $g_{\mathcal{B}}(\cdot): \mathbb{R} \rightarrow [0, 1]$ and assigns a drop probability to every unit $k$ in every layer $l$.
After bucketization, binary masks are sampled from a Bernoulli distribution for every layer $l$.

Since each neuron has now varying drop probability, we need to approximate its ``mean" drop probability in Eq.~\eqref{rescale_trick}, to implement the rescaling trick. We use the exponential moving average as of:  
\begin{equation} \label{ema-trick}
    p_k^{(l)}[n] = 
    \begin{cases}
        p_0 &, n=0\\
        (1 - \alpha) p_k^{(l)}[n-1] + \alpha g_{\mathcal{B}}(\tilde{Y}_k^{(l)}[n]) &, n>0
    \end{cases}
\end{equation}
where $\alpha \in [0,1]$ is a tunable hyperparameter named \textit{elasticity}, $n$ is the update step index and $p_0$ the initial drop probability for all units. 
When $\alpha=0$, Eq.~\eqref{ema-trick} is reduced to the vanilla dropout rescaling trick, with $p=p_0$.
After performing buketization, the activations are masked, rescaled and the weights are updated through back-propagation (Phase 2 in Fig.~\ref{fig:Y-Drop}).
Our approach is layer-wise, meaning that it is applied separately to each fully connected layer.

In order to calculate the expected value of units to be dropped in every step, we introduce an additional quantity, the \textit{overall mean probability}, which is defined as:
\begin{equation}
    p_M = w_{\mathcal{S}} p_H + w_{\mathcal{W}} p_L 
\end{equation}
We refer to $p_M$ as overall mean probability, because it is the corresponding quantity to the dropout probability $p$ \footnote{In the original dropout paper \cite{hinton_improving_2012} the authors refer to keep probability. In this work we use the drop probability since this is the quantity used in modern deep learning frameworks.} in the vanilla algorithm.
Depending on the values of bucket probabilities two limit cases can be distinguished.
In the first $p_{H}=p_{L}$ and the algorithm is reduced to regular dropout. In the second $p_H = 1$, $p_L = 0$ and the algorithm becomes deterministic. 

\subsection{Algorithm Implementation}
Through experimentation, we found that using fixed bucket probabilities $p_L, p_H$ requires a lot of tuning and therefore we opt for a more efficient approach.
Additionally, prior works \cite{morerio2017curriculum, wang2019jumpout} discuss the potential drawbacks of using fixed dropout rates.
We follow \cite{wang2019jumpout} and sample the values for $p_L, p_H$ in every step from the right half of a Gaussian distribution.
Formally described $p_L \sim \mathcal{N}_R(\mu_L, \sigma_L)$ and
$p_H \sim \mathcal{N}_R(\mu_H, \sigma_H)$, where $\mu$ denotes the mean value and $\sigma$ the respective standard deviation of the corresponding distribution.
We denote as $\mathcal{N}_R$ the right half (decreasing) of the normal distribution. 
Moreover, in order to better control the resulting probabilities we truncate the normal distributions using some thresholds $p_H^{max}, p_L^{max}$ as shown in the following formula
\begin{equation}
    p_{b} = \min\{\mathcal{N}_R(\mu_b, \sigma_b), p_b^{max}\}
\end{equation}
where $p_b$ denotes the bucket probability, i.e strong or weak.
The potential reason for the effectiveness of this modification can be attributed to the use of varying dropout rates, 
\textit{i.e.} the number of dropped neurons is not fixed and thus the algorithm becomes more robust to the choice of this hyperparameter.


\begin{table}[t]
    \caption{Y-Drop hyperparameter values across all examined scenarios}
    \vskip 0.35pt
	\centering
    \resizebox{0.6\columnwidth}{!}{
	\begin{tabular}{l c}
	    \toprule
	    \textbf{Hyperparameters} & \textbf{Default Value} \\
	    \midrule
		$n_c$ & 5 \\
		$w_{\mathcal{S}}, w_{\mathcal{W}}$ & 0.5 \\
		$\mu_L$ & 0.1 \\
		$\mu_H$ & 0.6 \\
		$\sigma_L, \sigma_H$ & 0.05 \\
		$p_0$ & 0.35 \\
		\bottomrule
	\end{tabular}
	}
\label{tab:hyperparams}
\end{table}

\par

During the first training steps neurons are randomly initialized and do not have meaningful conductance values, therefore it is hard to apply Y-Drop from the start of the training. This is known as the cold-start problem.
To address it, we use vanilla dropout at the start of the training and after (few) $\mathcal{K}$ epochs we switch to Y-Drop. 
We refer to this hyperparameter as \textit{annealing factor}. 
Similar schedules have been incorporated in other dropout variants, e.g. \cite{zoph2018learning, ghiasi_dropblock_2018, zeng_corrdrop_2020}
\par
As discussed in Section~\ref{sec:exp_setup} we only tune the elasticity $\alpha$ and the annealing factor $\mathcal{K}$. We fix the rest of the hyperparameters to the values shown in Table~\ref{tab:hyperparams}.

\subsection{Optimizing Memory Requirements} \label{mem_optimization}

Conductance calculation, as shown in Eq.~\eqref{cond_2}, needs an effective batch size of $n_c B$, where $n_c$ are the interpolation steps and $B$ the batch size.
We impose a fixed memory restriction so that Y-Drop does not require more memory than the original dropout.
To meet this restriction randomly select $B_c$ of the $B$ samples in every batch to use for the conductance calculation. $B_c$ is selected according to Eq.~\eqref{eq:mem_restriction}:
\begin{equation} \label{eq:mem_restriction}
    n_c B_c \leq B
\end{equation}
This inequality informs us on the number of samples we are allowed to use in order to calculate conductance for a fixed value of integration steps. 
This approximation does not affect Y-Drop's performance, since we are only interested in the ordering of the neurons according to their conductance, rather than exact conductance values.
We use $n_c=5$ integration steps in all our experiments (see Table~\ref{tab:hyperparams}).

\begin{table*}[!htb]
\centering
\caption{Comparative results for varying network sizes on MNIST and CIFAR-10. The $FC$ denotes fully connected architectures, while $S$ a CNN followed by different FC parts. The \textit{Params} column depicts the number of trainable parameter of the fully connected parts for every architecture. The last column shows the absolute improvement of Y-Drop over Dropout.}
\vskip 0.15in
    \centering
    \resizebox{0.7\linewidth}{!}{
    \begin{tabular}{lccccccc}
      \toprule 
      Dataset & Model & Params (M)
        & Plain
        & Dropout
        & Y-Drop 
        & $\Delta_Y -\Delta_{Drop}$\\
      \midrule 
      \multirow{4}{*}{MNIST} & $FC_1$ &$2$ & $98.18 \pm 0.05$ & $98.30 \pm 0.05$  & $98.47 \pm 0.10$ & $\mathbf{0.17}$ \\
      &$FC_2$ &$10$ & $98.19 \pm 0.16$ & $98.21 \pm 0.10$ & $98.53 \pm 0.13$ & $\mathbf{0.32}$\\
      &$FC_3$ &$54$ & $97.99 \pm 0.23$ & $98.18 \pm 0.16$ & $98.54 \pm 0.06$ & $\mathbf{0.36}$\\
      &$FC_4$ &$208$ & $97.99 \pm 0.27$ & $98.20 \pm 0.08$ & $98.58 \pm 0.06$ & $\mathbf{0.38}$ \\
      \bottomrule
    \end{tabular}
    }
    \label{tab:MNIST_scalability}
    \vskip 15pt
    \centering
    \resizebox{0.7\linewidth}{!}{
    \begin{tabular}{lcccccc}
      \toprule 
     Dataset & Model & Params (M)
        & Plain
        & Dropout
        & Y-Drop 
        & $\Delta_Y -\Delta_{Drop}$ \\
      \midrule 
      \multirow{4}{*}{CIFAR-10} & $S_1$ & $2$ & $81.15 \pm 0.24$ & $83.39 \pm 0.21$  & $83.77 \pm 0.18$ & $\mathbf{0.38}$ \\
      &$S_2$ & $10$& $81.69 \pm 0.20$ & $83.55 \pm 0.26$ & $84.10 \pm 0.27$ & $\mathbf{0.55}$ \\
      &$S_3$ & $54$& $81.63 \pm 0.20$ & $83.38 \pm 0.14$ & $84.20 \pm 0.24$ & $\mathbf{0.82}$ \\
      &$S_4$ & $87$ & $81.59 \pm 0.10$ & $83.47 \pm 0.33$ & $84.32 \pm 0.35$ & $\mathbf{0.85}$ \\
      \bottomrule
    \end{tabular}
    }
    \label{tab:CIFAR-10_scalability}
  \label{tab:scalability}
\end{table*}

\begin{table*}[!htb]
\caption{Comparative results of Y-Drop. The $\Delta$ quantities denote the absolute improvement over the baseline (Plain) for each method. The architectures $M1$ and $M2$ are CNNs followed by FC layers. The regularizers are applied to the FC part of the network.}
\label{tab:Ydrop-Result-I}
\vskip 0.15in
  \begin{center}
  \resizebox{0.9\linewidth}{!}{
    \begin{tabular}{lccccccccc}
      \toprule 
      \textbf{Dataset} 
        & \multicolumn{2}{c}{\textbf{STL-10}} 
        && \multicolumn{2}{c}{\textbf{CIFAR-100}} 
        && \multicolumn{2}{c}{\textbf{SVHN}} \\
        \cmidrule{2-3} 
        \cmidrule{5-6}
        \cmidrule{8-9}
      & $M_1$ & $M_2$ 
      && $M_1$ & $M_2$ 
      && $M_1$ & $M_2$\\
      \midrule 
      Plain
      & $72.51 \pm 0.32$ & $71.65 \pm 0.48$
      && $57.43 \pm 0.32$ & $56.80 \pm 0.15$
      && $94.31 \pm 0.10$ & $94.42 \pm 0.06$\\
      Dropout
      & $74.14 \pm 0.31$ & $73.55 \pm 0.23$
      && $61.97 \pm 0.24$ & $61.97 \pm 0.16$
      && $94.45 \pm 0.07$ & $94.38 \pm 0.09$\\
      Y-Drop
      & $\mathbf{74.68 \pm 0.42}$ & $\mathbf{74.26 \pm 0.26}$
      && $\mathbf{62.90 \pm 0.26}$ & $\mathbf{63.18 \pm 0.32}$
      && $\mathbf{94.61 \pm 0.05}$ & $\mathbf{94.82 \pm 0.08}$\\
      \midrule
      $\Delta_{Drop}$
      & $1.63$ & $1.90$
      && $4.54$ & $5.17$
      && $0.16$ & $-0.04$ \\
      $\Delta_{Y}$
      & $2.17$ & $2.61$
      && $5.47$ & $6.38$
      && $0.3$ & $0.4$ \\
      \bottomrule
    \end{tabular}
    }
  \end{center}
\end{table*}


\section{Experimental Setup} \label{sec:exp_setup}
\noindent \textbf{Datasets.} We verify the effectiveness of the proposed regularization algorithm on the following benchmark datasets. The \textit{MNIST} handwritten digit classification task \cite{lecun1998gradient}, which consists of $28\times 28$ black and white images, each containing a digit $0$ to $9$. The dataset contains $60,000$ training images and $10,000$ test images. The \textit{CIFAR-10} and \textit{CIFAR-100} \cite{krizhevsky2009learning} which are natural $32 \times 32$ RGB image datasets. The first contains $10$ classes and $50,000$ images for training and $10,000$ images for testing. The later has $100$ classes with $500$ training images and $100$ testing images per class. We also use the Street View House Numbers (\textit{SVHN}) dataset \cite{netzer2011}, which includes $604,388$ training images and $26,032$ testing images of size $32\times 32$. Also \textit{STL-10} \cite{coates2011analysis} (10 classes) which contains $96 \times 96$ RGB images with $500$ training samples and $800$ testing samples per class, is used.
\par
\noindent \textbf{Architectures.} The architectures we experiment with are denoted as $FC$, $S$ and $M$.
$FC$ describes a feedforward network, that consists of L layers with H units per layer with ReLU activation \cite{nair2010rectified}. A final layer is used for classification. Specifically, we train the following architectures $FC_1: 2\times 1024$, $FC_2: 3 \times 2048$, $FC_3: 4 \times 4096$, $FC_4: 4 \times 8192$. This architecture is used for MNIST classification.
$S$ is a CNN\footnote{\href{https://code.google.com/archive/p/cuda-convnet/}{layers80sec.cfg }: $3$ layer CNN with $32$, $32$ and $64$ filters respectively} proposed in \cite{krizhevsky2009learning}, followed by $L$ fully connected layers with $H$ units and ReLU activations. Again we vary the number and size of fully connected layers in four configurations, i.e. $S_1: 2 \times 1024$, $S_2: 3\times 2048$, $S_3: 4\times 4096$ and $S_4: 6 \times 4096$. The convolutional layers stay the same for $S_1, S_2, S_3, S_4$. This architecture is used for CIFAR-10.
For STL-10, SVHN and CIFAR-100 we use  a CNN which has $3$ layers with $96, 128$ and $256$ filters respectively \cite{krizhevsky2009learning}, denoted as $M$. Each convolutional layer has a $5\times 5$ receptive field with a stride of $1$ pixel. For the max pooling layers we choose $3 \times 3$ regions with stride $2$. We employ Batch normalization \cite{ioffe2015batch} and use ReLU activation.
We build the $M_1$ architecture by attaching 2 fully connected layers, with $2048$ units each, on top of the 3-layered CNN.
Similarly, we build $M_2$ with adding 3 fully connected layers with $4096$ neurons each. We apply Y-Drop only to the fully connected layers.
\par
\noindent \textbf{Setup.} All models are implemented using \texttt{PyTorch} \cite{paszke2019pytorch} and trained using SGD with momentum ($\beta=0.9$). For $M_1$ and $M_2$ a step learning rate scheduler with initial learning rate of $0.01$ is used.
Moreover in CIFAR-10, CIFAR-100 and STL-10 standard data augmentation, i.e random flipping and cropping, is used. Vanilla dropout parameter is tuned in the range $[0.1, 0.6]$. We consistently found that $0.5$ is the best performing value in all setups. All experiments are carried out with constant batch size $B=64$, except from STL-10 and SVHN where batches of size $32$ are used. 
We use $25\%$ of the training set for hyperparameter validation. For STL-10 we use $10 \%$ of the training set for validation. We use early stopping on the validation loss with patience of 10 epochs.
All reported results are averaged over $5$ runs.
\par \noindent \textbf{Y-Drop tuning.}
The elasticity value $\alpha$ is tuned to the optimal value from the set $\{10^{-5}, 10^{-4}, 10^{-3}, 10^{-2}, 10^{-1}\}$.
We also tune the annealing factor $\mathcal{K}$ in the range $[1,10]$ with step 2.
Conductance is calculated using \texttt{Captum} \cite{kokhlikyan2020captum}.
Following Section~\ref{mem_optimization} we set $n_c=5$.
Subsequently, $12$ and $6$ samples are used to calculate conductance for batch sizes $64$ and $32$ respectively. All other hyperparameters are set as in
Table~\ref{tab:hyperparams} for all experiments.

\begin{figure*}[!htb]
     \centering
     \begin{subfigure}[b]{0.48\textwidth}
         \centering
        \includegraphics[width=\textwidth]{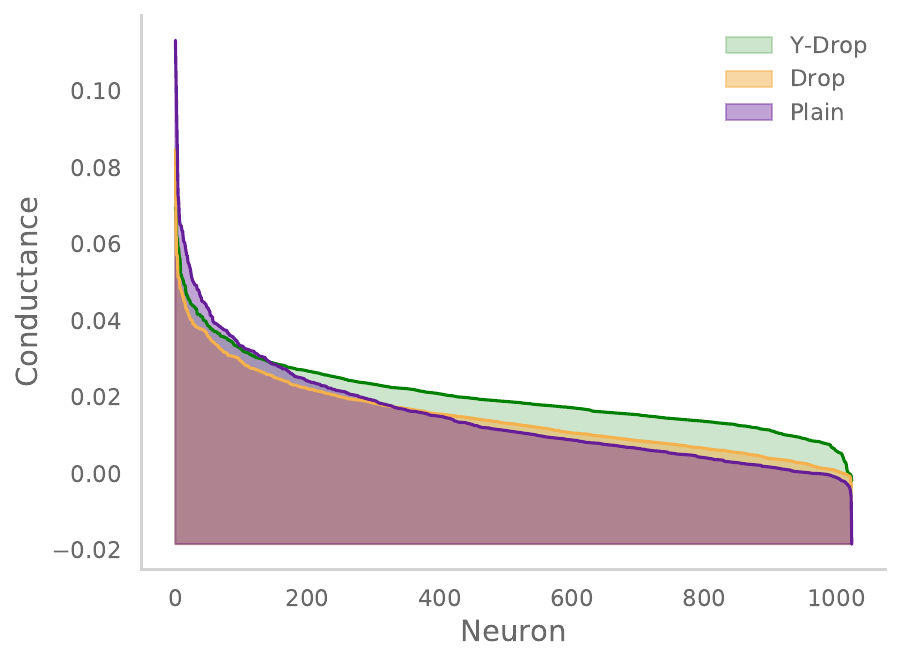}
        \caption{}
         \label{fig:mean-cond}
     \end{subfigure}
     \hfill
     \begin{subfigure}[b]{0.48\textwidth}
         \centering
         \includegraphics[width=\textwidth]{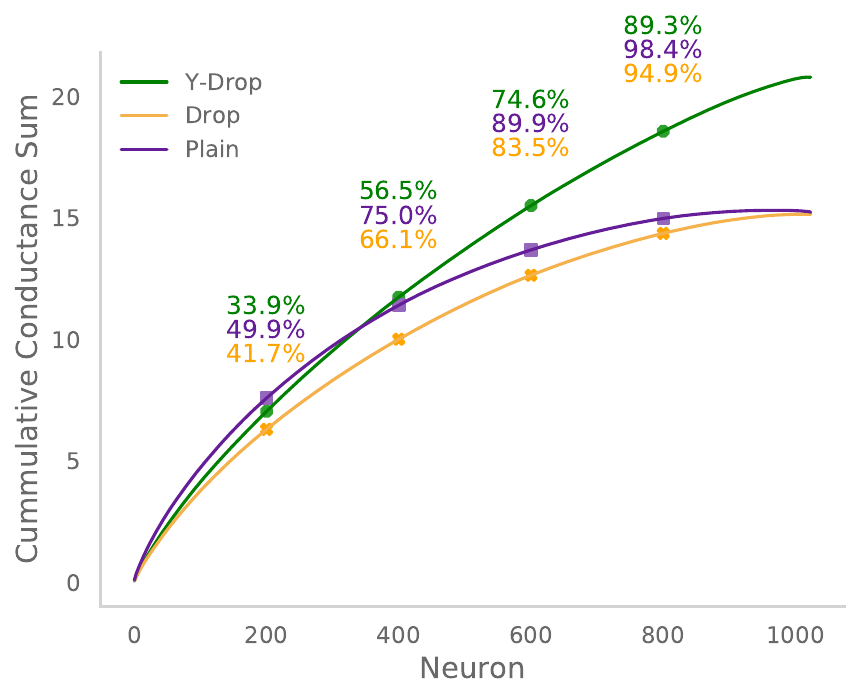}
        \caption{}
         \label{fig:cumsum-cond}
     \end{subfigure}
     \caption{
     Illustration of the \textit{average neuron conductance scores} for $1024$ units in a single layered network trained on MNIST, using Y-Drop (green), Dropout (orange) and no regularization / Plain (purple).
     Fig.~\ref{fig:mean-cond} shows the mean neuron conductance of the three models. 
     Fig.~\ref{fig:cumsum-cond} shows the cumulative sum of conductance over units.
     Colored numbers in Fig.~\ref{fig:cumsum-cond} indicate the percentage of the total layer conductance when the top $200, 400, 600$ or $800$ units are taken into consideration.
     Units in both Figures are sorted from highest conductance score to lowest. 
     }
     \label{fig:mean_cumsum_conductance}
\end{figure*}

\begin{figure*}[!tbh]
     \centering
     \begin{subfigure}[b]{0.48\textwidth}
         \centering
         \includegraphics[width=\textwidth]{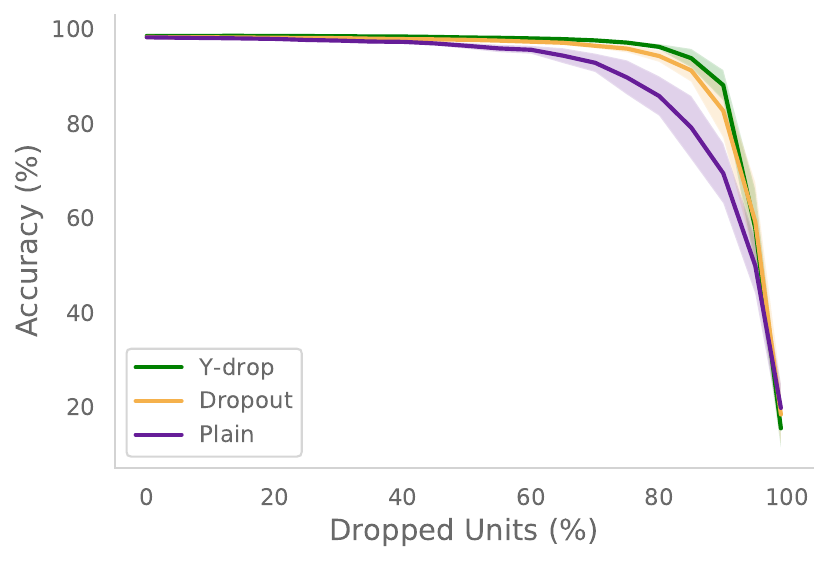}
        \caption{}
        \label{fig:random_mask}
     \end{subfigure}
     \hfill
     \begin{subfigure}[b]{0.48\textwidth}
         \centering
         \includegraphics[width=\textwidth]{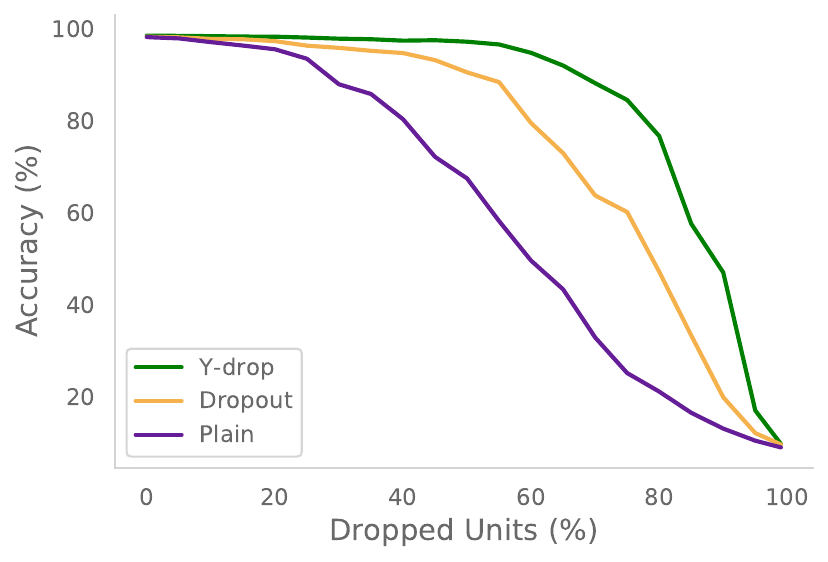}
        \caption{}
        \label{fig:cond_mask}
     \end{subfigure}
     \caption{Performance of Y-Drop (green), Dropout (orange) and no regularization / Plain (purple) trained networks, when we drop progressively higher percentage of units (during inference).
     In Fig.~\ref{fig:random_mask} we drop units randomly.
     In Fig.~\ref{fig:cond_mask} we first drop units with higher conductance scores.
     Both figures end when $99\%$ of units are dropped.}
     \label{fig:mask_out_experiments}
\end{figure*}

\section{Experiments}
\subsection{Comparison with Dropout}
In Table~\ref{tab:scalability} we perform experiments with varying network sizes to study the regularization ability of Y-drop.
Specifically we perform classification on MNIST using the feedforward architectures $FC_1$, $FC_2$, $FC_3$, $FC_4$, descripted in Section~\ref{sec:exp_setup}.
We see that Y-Drop consistently improves performance over dropout.
Interestingly, Y-Drop performance improvement gets larger as the architecture size increases. Note that for the largest architectures $FC_3$ and $FC_4$ dropout performance degrades, while Y-drop performance increases with the number of parameters.
Note that for Guided Dropout \cite{keshari2019guided}, which is another dropout variant, performance on a similar experiment degrades for larger networks.
We repeat this experiment for CIFAR-10 using the convolutional $S1$, $S2$, $S3$ and $S4$ architectures and observe similar behavior. This showcases that Y-drop is able to regularize effectively even extremely overparameterized configurations.

In Table~\ref{tab:Ydrop-Result-I} we compare the performance of Y-Drop and Dropout for the larger convolutional networks $M_1$ and $M_2$ for STL-10, CIFAR-100 and SVHN. $\Delta_Y$ and $\Delta_{Drop}$ denote the absolute improvement over the unregularized (Plain) network for Y-drop and dropout respectively.
We observe that Y-drop consistently yields larger performance improvements over the unregularized network compared to dropout.
The examined scenarios vary in the amount of training data, complexity and number of classes.
We observe that for the smaller datasets, i.e. STL-10 and CIFAR-100, Y-drop yields larger performance improvements compared to vanilla dropout, again indicating stronger regularization.
On top of that, the experiments on CIFAR-100 empirically verify that Y-Drop is effective even for scenarios where the number of classes exceeds the batch size and we are not able to capture information for all classes at every iteration step for conductance calculation.

\par

\subsection{Y-drop spreads conductance across more neurons}
We perform an analysis of Y-Drop by computing the \textit{average conductance per neuron} in Fig.~\ref{fig:mean-cond}. 
For this experiment we perform classification on MNIST using $1$ fully connected layer with $1024$ units with ReLU activation, followed by a classification layer.
We compare conductance scores for this architecture, when trained without regularization (Plain), with dropout (Drop) and with Y-Drop. Mean conductance scores are calculated on the validation set.
Neurons are sorted from most important to least important.
We observe that the conductance scores are more spread across units when using Y-Drop, namely more neurons are important and contribute to the classification.
For dropout and Plain setups, we observe that $~20\%$ of the neurons have very high conductance scores (higher than Y-Drop), but the rest of the neurons have low conductance.

This is more clearly illustrated in Fig.~\ref{fig:cumsum-cond}, where we plot the cumulative sums of conductance for the $1024$ neurons of the aforementioned architectures. We see that the network trained using Y-Drop has $~37\%$ higher overall conductance than dropout and Plain setups.
In Fig. \ref{fig:cumsum-cond} we can also see the conductance percentiles at each point, i.e. the percentage of the total conductance accumulated by the neurons at this point.
We see that only $20 \%$ of all neurons 
contribute $50\%$ of the total network conductance for the unregularized network and $41.7\%$ for the network trained using dropout. For Y-Drop we see an almost linear increase in the conductance percentage contributed as we increase the number of units.

Y-Drop results in networks with more important neurons distributed across the architecture. This allows the network to use more of its capacity to solve the task at hand, which in general does not apply to other methods \cite{dauphin2013}. We expect these networks to demonstrate greater robustness and be less prone to overfitting.

\subsection{Reliance on important units}

\citet{morcos2018importance} show that reliance on specific, or limited, neurons is an overfitting indicator. To further investigate whether better distributed neuron conductance results in more robust networks we carry two additional experiments. We employ our analysis on networks trained with no regularization, i.e. Plain, with Dropout and Y-Drop. 

For all three networks we show the performance degradation when dropping units during inference, i.e. prune units and evaluate.
In Fig. \ref{fig:random_mask} we randomly drop progressively higher percentages of units (during inference), for the Plain, Dropout and Y-Drop networks. Accuracy scores are averaged over $25$ runs and we show mean and standard deviation in Fig.~\ref{fig:random_mask}. We see that dropout and Y-Drop are more resilient than the unregularized (Plain) network, with Y-Drop being slightly more robust. In Fig.~\ref{fig:cond_mask} we follow a more aggressive strategy, by dropping units with higher conductance first. Again conductance scores are calculated on the validation set. We see that the Plain network performance starts decreasing even when only $20\%$ of the most important neurons are dropped. The network trained using dropout is robust up to $60\%$ of dropped units. Observe that Y-Drop performance starts rapidly decreasing after $80\%$ of the neurons are dropped. 
This indicates that nets trained with Y-drop are more robust due to more units contributing to solve the task at hand.
This built-in redundancy can be another step towards avoiding neuron co-adaptation and improving network generalization ability, which is the intended purpose of the original dropout algorithm.


\section{Conclusions}

In this work, we indroduce Y-Drop, a regularization algorithm that integrates neural conductance into dropout during network training.
Conductance is an interpretability measure that assigns higher scores to more important units wrt the network prediction.
The proposed algorithm uses conductance to drop more important units with higher probability, forcing the network to solve the task at hand using weaker neurons.
In our experiments we show that this approach provides strong regularization, yielding consistent improvements across five datasets.
Our approach scales with the size of the architecture and is able to regularize even highly over-parameterized networks.
Our analysis shows that networks trained with Y-drop 
have more units with high conductance scores and subsequently rely less on a small amount important units. Y-drop is easy to tune and adapt for new tasks.

In the future, we plan to extend Y-drop for other architectures, i.e. CNNs and RNNs and apply it to more diverse problem settings, aiming at a universal drop-in replacement for dropout.
Additionally, we plan to integrate other importance measures, such as Internal Influence or GradCAM and investigate their regularization properties.
Finally we will investigate whether efficient pruning techniques can be developed using importance-based dropout.

\nocite{langley00}

\bibliography{example_paper}
\bibliographystyle{icml2021}





\end{document}